\documentclass[10pt,twocolumn,letterpaper]{article}

\usepackage{wacv}
\usepackage{times}
\usepackage{epsfig}
\usepackage{graphicx}
\usepackage{amsmath}
\usepackage{amssymb}
\usepackage{booktabs}
\usepackage{float}

\def\wacvPaperID{1592} 

\wacvalgorithmstrack   

\wacvfinalcopy 

\ifwacvfinal
\usepackage[breaklinks=true,bookmarks=false]{hyperref}
\else
\usepackage[pagebackref=true,breaklinks=true,colorlinks,bookmarks=false]{hyperref}
\fi

\begin{document}

\title{FUSSL: Fuzzy Uncertain Self Supervised Learning}

\author{Salman Mohamadi\\
West Virginia University\\
Morgantown, WV, USA\\
{\tt\small sm0224@mix.wvu.edu}
\and
Gianfranco Doretto\\
West Virginia University\\
Morgantown, WV, USA\\
{\tt\small gianfranco.doretto@mail.wvu.edu}
\and
Donald A. Adjeroh\\
West Virginia University\\
Morgantown, WV, USA\\
{\tt\small donald.adjeroh@mail.wvu.edu}
}

\maketitle
\thispagestyle{empty}

\begin{abstract}
Self supervised learning (SSL) 
has become a very successful technique  to
harness the power of unlabeled data, with no 
annotation effort. A number of developed approaches are evolving with the goal of outperforming supervised alternatives, which have been relatively successful. Similar to some other disciplines in deep representation learning, one main issue in SSL  
is robustness of the approaches under different settings. 
In this paper, for the first time, we recognise the fundamental limits of SSL coming from the use of a single-supervisory signal. 
To address this limitation, we leverage the power of uncertainty representation to devise a robust and general standard hierarchical learning/training protocol for any SSL baseline, regardless of their assumptions and approaches. Essentially, using  the information bottleneck principle, we decompose feature learning into a two-stage training procedure, each with a distinct supervision signal. This double supervision approach is captured in two key steps: 1) invariance enforcement to data augmentation, and 2) fuzzy pseudo labeling (both hard and soft annotation). This simple, yet, effective protocol which enables cross-class/cluster feature learning, is instantiated via an initial training of an ensemble of models through invariance enforcement to data augmentation as first training phase, and then assigning fuzzy labels to the original samples for the second training phase. 
We consider multiple alternative scenarios with double supervision and evaluate 
the effectiveness of our approach on recent baselines, covering four different SSL paradigms, including geometrical, contrastive, non-contrastive, and hard/soft whitening (redundancy reduction) baselines. We performed extensive experiments under multiple settings to show that the proposed 
training protocol consistently improves the performance of the former baselines, independent of their respective underlying principles.

\end{abstract}
\vspace{-.8cm}
\section{Introduction}
\vspace{-.21cm}
Self supervised learning typically involves the use of a pretext task and an objective (loss) function, whereby a supervisory signal from the unlabeled data is used to learn an appropriate representation, based on the objective function. Even though its goal is somehow similar to those of deep active learning \cite{mohamadi2020deep,mohamadi2022deep} or semi-supervised learning, here the specific goal is primarily to learn a range of general features from the large amounts of unlabeled data via a proxy task. The learnt features can then be  applied for improved performance on a downstream task, such as semantic segmentation, object detection, and image captioning \cite{ermolov2021whitening,jing2020self}.
For some applications, such large amounts of unlabelled data are often available at a minimal cost. 
Though one might trace the inception of SSL back to few decades ago in work such as \cite{becker1992self}, the modern view 
of SSL in form of decomposing it into two integral components (pretext task and loss function) was not until recent years. This idea of learning a pretext task as a proxy task via optimising a proper loss (objective) function has evolved in both directions: new pretext tasks as well as improved loss functions. More specifically, in computer vision, a number of pretext tasks have been 
proposed, e.g.,  view discriminative tasks for image data, and temporal consistency or temporal cycle-consistency for video data, each requiring its own idiosyncratic architectural design. However, recent work \cite{chen2020simple} proposed a framework needless of pretext task specific network design. Loss functions have also evolved, though in a less diversified manner. From a reductionist view, the general idea of almost all of the loss functions in SSL is to enforce invariance to the representation of perturbed data 
at the sample level or cluster level. 
At sample level, for a given sample $x$, two (or more) augmented views will be generated via a random augmentation process $\tau$, and depending on the type of loss function, these views (also know as positive views) will be contrasted only against each other or against each other and augmented views from other samples (negative views). Specifically, for an image sample $x$, let  $x_i$ and $x_j$ be  two positive examples while $x_k$ is a negative example, each  with respective latent space representations $z_i$, $z_j$, and $z_k$. Generally speaking, earlier loss functions attempted to map positive examples as close as possible in a latent space, while discriminating against negative examples via a contrastive loss \cite{chen2020simple}, though with the possibility of representation collapse. Latter loss functions known as non-contrastive approaches \cite{grill2020bootstrap,chen2021exploring}, eliminated the need for negative pair contrast while delivering higher performance, with little or no risk of representation collapse. Most recently, two negative-free methods \cite{ermolov2021whitening,zbontar2021barlow} are proposed which perform hard or soft whitening to reduce the redundancy in representation. There is also another set of approaches known as clustering (geometrical) approaches such as \cite{caron2020unsupervised} which utilize a fundamentally different supervisory signal, eliminating the effect of data augmentation in geometrical representation.

Even though both major components of SSL frameworks, pretext task and loss function, are noticeably diversified, we 
observe that this has not been the case for the supervisory signal.
Researchers turned to SSL to eliminate the need for supervised data annotation, by replacing the supervision with data-derived supervisory signals. These signals include implicit binary labeling for sample level invariance enforcement as well as geometry-based (clustering-based) invariance enforcement to the data augmentation 
However, it turned out that each type of supervisory signal comes with its downsides, due to the fact that they mainly  \textbf{target only a certain level of granularity with respect to the features}.


More specifically, on one hand, rigorous invariance enforcement to data augmentation at the sample level leads to a slight deterioration in performance over the downstream task \cite{huang2022learning}. On the other hand, cluster-based supervisory signals are weak at generalization, and often might under-perform in case of stiff transfer learning. The problem with former supervisory signals is implicitly addressed in \cite{oord2018representation,wang2021dense,xie2021propagate,roh2021spatially,xiao2021region} mostly by using downstream task-specific solutions, or somehow modified pretext tasks \cite{huang2022learning}. Unlike these often non-generalizable solutions, we aim for a general solution regardless of downstream task or pretext task. The problem with cluster-based 
supervisory signals seems to be less recognized. We  suspect this to be mainly due to the fact that cluster-based approaches tend to deliver relatively more robust and better performance as they target learning higher granularity of features (cluster level features as opposed to sample level).
We consider the above as a joint problem from the lens of information theory to come up with a general solution. Hence, in this paper we explicitly address the issue by devising a general standard training protocol and evaluate it on a number of recent  baselines. The prime idea of the work is hierarchical training via double-supervision. In fact, we integrate two phases of training built upon (I) low level feature learning via sample-level invariant representation enforcement  (phase 1), and (II) medium to high level feature learning via cluster-level representation learning using training over uncertainty-driven pseudo labeled data (phase 2). 
That is, in phase 2  we perform pseudo labeling in a fuzzy setting (both hard and soft labels) right after the first phase of training.
In essence, we show this training paradigm improves each and every former baseline by incorporation of high and low level feature learning which cancels out the effect of rigorous sample-level invariance enforcement as well as cluter-level non-generalizability. This paper has three main  contributions:
\begin{itemize}
    \item We propose a new standard training protocol for SSL frameworks based on double-supervisory signals, which enjoys hierarchical feature learning via both sample-level invariant representation as well as cluster-level learning using training on soft pseudo labeled data. This standard protocol is applicable to all former baselines and future baselines, eliminating performance drop on downstream tasks due to the use of single-supervisory signal.
    \item We exploit the benefit of uncertainty representation using a new architectural design to enhance the robustness of several recent baselines trained via our new training protocol. Furthermore, this protocol neither suffers from the downside of rigorous sample-level invariance enforcement nor the computational overhead of clustering-based approaches.
    \item We perform extensive experiments to show the effectiveness of the proposed training protocol as well as analyse multiple scenarios to evaluate the impact of different algorithmic parameters on the performance.
\end{itemize}
\vspace{-.5cm}
\section{Preliminary and background}
\vspace{-.21cm}
Pretext task and loss function are integral components of any SSL framework enabling a self supervision signal. Hence in this section we present the evolution of loss functions from the perspective of supervisory signal, and also explore the uncertainty representation in the context of SSL.
\vspace{-.15cm}
\subsection{Loss functions}
\vspace{-.21cm}
Below, we briefly discuss several objective functions, including triplet loss, typical contrastive loss, and recent non-contrastive loss functions, all sharing the same approach to supervisory signal, sample-level supervision. Then we recognize another type of supervisory signal embedded in clustering-based losses, cluster-level supervision. 
\vspace{-.21cm}
\subsubsection{Sample-level supervision}
\vspace{-.15cm}
\textbf{Triplet loss:} Triplet loss is a discriminative loss \cite{misra2016shuffle,wang2015unsupervised}, in which given three latent spaces $z_j$, $z_j$ and $z_k$, this loss explicitly aims for minimizing the distance between positive pairs ($z_i$ and $z_j$); and maximizing the distance between negative pairs ($z_i$ and $z_k$) as presented below:
\begin{equation}
    \label{eq2}
    \mathcal{L}_{\bigtriangleup}=\max(0, z_i^T z_j- z_i^T z_k+m),
\end{equation}
with $m$ as a margin hyperparameter.

Also multi-class N-pair as a generalization of triplet loss for joint comparison among more than one negative example was developed by \cite{sohn2016improved} as follows:
\begin{equation}
    \label{eq1}
  \mathcal{L}_{z_i,z_J} = \log\left( 1+\sum_{k=1,k\neq i}^{2N}\exp(z_i z_k - z_i z_j)\right)
\end{equation}
\textbf{Contrastive loss:} 
A popular discriminative loss in SSL up until recently \cite{oord2018representation,tian2020contrastive,he2020momentum,chen2020simple,bachman2019learning}, was contrastive loss, which is a very demanding loss in terms of required batch size of negative instances to reduce the risk of its representation collapse. Updated versions of this loss are still demanding either computationally or in terms of negative batch size. Wang and Gupta \cite{wang2015unsupervised} reformulated the basic contrastive loss for SSL with $N-1$ negative examples and $\tau$ as a temperature hyperparameter as follows:
\begin{equation}
    \label{eq3}
    \mathcal{L}_{Contrastive} = -\log \frac{\exp (z_i^T z_j/\tau)}{\sum_{n=1, n\neq i}^{N}\exp (z_i^T z_k/\tau)}.
\end{equation}

\textbf{Non-contrastive loss functions:} BYOL \cite{grill2020bootstrap} devised a type of loss 
without using 
negative instances while avoiding representation collapse, making a key contribution in 
the use of non-contrsative loss. It was later followed by SimSiam \cite{chen2021exploring}. Tian et al. \cite{tian2021understanding} investigated the elements of these negative-free approaches relying on architectural update (adding a predictor block) as well as new training protocol (stop-gradient policy), which enables them to substantially outperform contrastive approaches while avoiding the trivial representation. 

Along this line of work on negative-free approaches, two recent approaches know as whitening-MSE and Barlow Twins \cite{ermolov2021whitening,zbontar2021barlow} primarily based on whitening the embedding or batches of embedding space established new baselines. Whitening-MSE also called hard whitening \cite{ermolov2021whitening} applies Cholesky decomposition to perform whitening over embeddings of a pair of networks, followed by a cosine similarity between the output of operations from each  of the networks as follows:
\begin{equation}
    \label{eq4}
    \min_{\theta} \mathbb{E}[2-2\frac{\langle z_i,z_j \rangle}{\lVert  z_i  \rVert_2 . \lVert  z_j \rVert_2}],\quad s.t.\quad cov(z_i,z_i)=cov(z_j,z_j)=I.
\end{equation}
{which ends up using MSE}

Barlow Twins \cite{zbontar2021barlow} also called soft whitening performs whitening over the square matrix cross-correlation of twin network outputs, $C$, which has relatively simpler loss function as follows:
\begin{equation}
    \label{eq6}
     \quad \mathcal{L}_{BT} \triangleq \sum_{i}(1-C_{ii})^2 + \lambda\sum_{i}\sum_{j\neq i}(C_{ij})^2, \:\:\: 
\end{equation}
\begin{equation}
    \label{eq6-2}
    C_{ij}\triangleq \frac{\sum_m z_{m,i}^{A} z_{m,j}^{B}}{\sqrt{\sum_m (z_{m,i}^{A})^2}\sqrt{\sum_m (z_{m,j}^{B})^2}}
\end{equation}
where $\lambda>0$ is a trade-off constant with typical value of $10^{-2}$, $m$ goes over batch samples, $i, j$ are indices bounded by the dimension of the outputs of the networks and $-1\leq C_{ij}\leq1$.

An interesting investigation by Balestriero and LeCun \cite{balestriero2022contrastive}, suggests that non-contrastive loss functions are generally more preferable due to better error bound on downstream tasks.
\vspace{-.21cm}
\subsubsection{Cluster-level supervision}
Unlike the above  methods which all directly involve features, a different set of approaches based on clustering \cite{caron2020unsupervised,caron2018deep,caron2021emerging} have evolved, primarily using cross-entropy loss in a geometric setting. They also enforce invariance to augmented instances but in a cluster setting rather than single-sample setting . We differentiate them from the former approaches in terms of the type of supervisory signal.\\
\textbf{Clustering-based loss function:} These are approaches based on clustering \cite{bautista2016cliquecnn,caron2020unsupervised,caron2018deep,caron2021emerging,ji2019invariant} which involve sample space representation. Basically, these primarily use cross-entropy loss in a geometrical setting to assign a cluster to samples targeting the semantic class representation rather than single sample. In terms of loss functions, for each original sample, a pair of augmented views is generated, in which one is used to guide the loss function to find a target and the other one aims at predicting the same target. This is generally formulated in the framework 
of  geometrical optimisation.
An interesting point about the clustering-based approaches is that they are also negative-free, similar to non-contrastive approaches. However, they  do not guarantee avoidance of the degenerate solution (representation collapse) and also incur computational overhead  due to the clustering process.

One canonical example include SwAV \cite{caron2020unsupervised} in which multiple positives are used to accomplish sample to cluster allocation via a cross entropy loss optimisation.
\vspace{-.21cm}
\subsection{Uncertainty and SSL:} 
\vspace{-0.21cm}
LeCunn \cite{Lexpodcast} suggests that uncertainty modeling in deep learning, and specifically in SSL is under-explored which would attract significant attention in the next decade (due to its immediate effectiveness). Model uncertainty in SSL is really under-explored, except notably in a few recent work such as \cite{hendrycks2019using,poggi2020uncertainty,liu2019exploiting}. These mainly used SSL for model uncertainty estimation as well as robustness improvement rather than improving SSL by that. For instance, Hendrycks \cite{hendrycks2019using} specifically weigh on the other beneficial aspect of SSL to enhance the performance on downstream task evaluation. Accordingly, they leverage SSL to improve model uncertainty and robustness such as handling adversarial instance and annotation corruption. Another work in \cite{poggi2020uncertainty} outlined the significance of accuracy in depth estimation and proposes uncertainty modeling to make the depth estimation more accurate.
With an emphasis on the concept of SSL in robotics and spatial perception, Nava et al \cite{nava2021uncertainty} proposed to apply uncertainty to former baselines in order to reduce the state estimation error.
\begin{figure}
\label{Fig1}
  \centering
  \includegraphics[scale=.27]{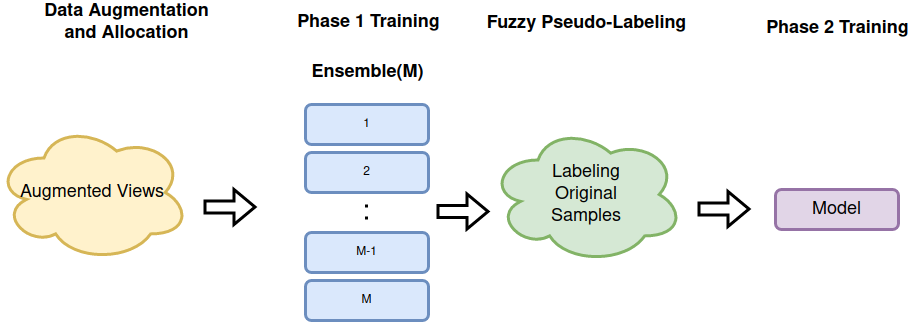}
		\caption{\scriptsize
		Schematic depiction of the proposed FUSSL architecture, to be built using any given SSL baseline; each block of the ensemble observes only its own set of augmented instances. Besides the sequential training, we also consider another scenario, namely progressive relabeling in the ablation study.
\vspace{-1.5em}
}
\end{figure}
Our work is in part inspired by \cite{lienen2021credal} which translates the concept of credal sets (set of probability distributions) to SSL in order to provide model uncertainty in pseudo-labels in low regime labeled data. This work proposes the use of credal sets to model uncertainty in pseudo-labels and hence reduce calibration errors in SSL approaches.\\
\vspace{-0.8cm}
\section{Hierarchical feature learning with FUSSL}
\vspace{-0.3cm}
\label{sec3intro}
Consider the downsides of single-supervisory signal SSL. As earlier discussed, sample-level and cluster-level supervisory signals come with their often serious downsides. Enforcing rigorous invariant representation at sample-level tends to be deleterious to some downstream tasks, whereas seeking invariant representation at cluster-level is computationally expensive and often non-adaptive 
to other data domains. Hence, we are motivated to devise a protocol that enjoys the benefit of both types of supervision while avoiding their respective downsides.
This protocol involves two phases of training. One guided by sample-level supervision to learn low to medium granularity features. The other uses cluster-level supervision to learn slightly higher level features by training over pseudo-labeled samples, which are predicted in a fuzzy setting, thanks to the first  training phase. 


If we view sample- and cluster-level approaches from the lens of information theory, both sets of approaches can be  shown to be reducible to a pair of terms which follow the Information Bottleneck (IB) principle \cite{tishby2000information, motiian2016information}.
Our general training protocol is defined based on decomposition of  the IB principle into \textbf{two pairs of terms} rather than one, where each pair represents a learning paradigm guided by a distinct supervisory signal. Specifically, in this context, the IB principle asserts that an SSL objective function learns a representation $Z$ which is invariant to random distortions applied to the sample, here denoted as $X$, while variant  to (or informative of) the sample distribution $Y$. Therefore, the goal is to have:
\begin{equation}
    \label{eq-IB}
    \centering
    \begin{split}
    & \min_{p(z|y)} IB=\min_{p(z|y)} (IB_s+IB_c) \\
    & \min_{p(z|y)} {IB}_{s \:\theta} \triangleq \min_{p(z|y)} \left( I(Z_\theta;Y)-\beta_s I(Z_\theta;X)\right)\\
    & \min_{p(z|y)} {IB}_{c \:\theta} \triangleq \min_{p(z|y)} \left( I(Z_\theta;Y)-\beta_c I(Z_\theta;X)\right)\\
    & \therefore  \text{we want:} \quad \min_{I{s,c}}\max_{IB_s, IB_c} \left(I_{sc}-IB_s -IB_c\right)
    \end{split}
\end{equation}
where $I(P;Q)$ denotes the mutual information between $P$ and $Q$, $IB_s$ represents learning guided by sample level supervisory signal, whereas $IB_c$ represents learning by cluster level supervision, and $I_{sc}=I(IB_s;IB_c)$. Now to maximize the total information extracted by $IB$, we need to maximize each one of  $IB_s$  and $IB_c$, while minimizing the mutual information (similar/identical features) between them. 
The minimization assures that the final representation does not collapse to the identical features learned by the two supervision steps, but to diversify as much as possible. It is worth mentioning that the worst case scenario happens when  both supervisions result in the an identical feature representation, and hence  the collective representation is as good as each of them. However, here we consider the element of time in terms of sequential learning, which is translated into implementation as a practical technique. Specifically, we decompose the learning into two sequential phases, where the learned features from the first phase would freeze slightly after initiating the second phase of training, in order to enact a conditional second phase of feature learning (conditioned on the previously learned features). This conditional learning enables a better exploration of the representation space (more diversified features), and consequently avoids the identical representation by both phases of training. Originally $IB_{s\:\theta}$ and $IB_{c\:\theta}$ in Eq. \ref{eq-IB} share the $\theta$, however, in our implementation we perform sequential training with a subset of fixed weights to avoid trivial representation. Hence, $IB_{c\:\theta}$ is eventually conditioned on learned weights from the phase 1 training, as follows: 
\begin{equation}
    \label{eq-IB}
    \centering
    \footnotesize
    \begin{split}
    & \min_{p(z|y)} (IB_{s,\theta_s}) \quad \text{followed by} \quad \min_{p(z|y,\theta_s)} (IB_{c,\theta_c})  \\
   & \therefore \min_{p(z|y,\theta_s)} {{IB}_{c,\theta_c}} \triangleq \min_{p(z|y,\theta_s)} \left( I(Z_{\theta_c};Y| {\theta_s})-\beta_c I(Z_{\theta_c}; X|{ \theta_{s}})\right)\\
    \end{split}
\end{equation}
\subsection{First phase training}
In the first phase of pre-training, an ensemble of size ${m}$ is built with ${m}$ building blocks. Each block is an independent model, with its own set of augmented examples as shown in Fig. 1 (Sec. \ref{Fig1}).
Data augmentation provides $2{m}$ distinct random augmented views for each sample, and allocate each pair of them to each block. In other words, depending on the size of the ensemble, a varying number of distorted examples would be generated randomly which allows the blocks to observe only their own set of examples for each original sample. That said, it is reasonably expected that for each sample, the total extracted information by ensemble to be more than each block alone. More formally, for sample $x$, there would be set of augmented views $\{(x_1,x_{1'}),(x_2,x_{2'}),...,(x_{{m}},x_{{{m}}'})\}$, where a given block ${i}$ with ${f}_b\circ {f}_p$ built from a backbone architecture ${f}_b$ and a projector ${f}_p$, to be trained on ${(x_i,x_{i'}) }$.
This accentuates the power of ensemble to search the representation space. Up until now in our framework, for a given baseline we only train ${m}$ copies of its model in an independent setting. 

There are elements of uncertainty representation both in the architecture as well as training examples, which later will facilitate fuzzy labeling of original samples for the second phase of feature leaning. To elaborate on architectural element of uncertainty, let's say an ensemble of identical models is trained on the same training samples. The representation of the data by each block will slightly differ due to the stochastic dynamics governing each model's parameters. On the other hand, another element of uncertainty comes from the augmentation; the input data for each block is a set of distinct randomly augmented examples, which even adds more uncertainty to the overall ensemble's representation of that data. Hence, the pipeline enjoys uncertainty from both data as well as model's parameters due to the architecture. Later we will see that the combination of these two can improve performance, regardless of the baselines used for building blocks, due to enriched robustness. 

The core learning idea of the first phase of training revolves around the supervisory signal inherent in the process of invariance enforcement to the positive augmented views. 
\vspace{-.5cm}
\subsection{Fuzzy pseudo labeling}
\vspace{-.15cm}
With the first phase of training both backbone ${f}_b$ and projector ${f}_p$ of each block learn low granularity general features. This comes to assist in the fuzzy labeling of the original samples with both hard labels as well as soft labels, i.e., ($0\leq$ label $\leq 1$). In fact, the original sample is fed to all blocks and the normalized output of the projector of each block assigns a label, where the final label either assigns one class to the sample or as many as ${m}$ classes, where $m$ is size of ensemble. If a class wins more blocks than others the ultimate label is a hard label assigning the sample to that class, otherwise the sample is assumed to be a member of multiple classes with different membership scores. In our ablation study we consider the counter-scenario of only having soft labels. Note that  the output dimension of all baselines assessed under our training protocol is 1000. As we do not have annotated data to better calibrate the output of all blocks that present the same label for the same hypothetical class (pseudo-class synchronized label assignment), we need to initialize all blocks with exactly identical initialization weights.
\vspace{-.21cm}
\subsection{Second phase training}
\vspace{-.15cm}
There is a second phase of training primarily distinguished by its supervisory signal. Unlike the former stage relying on invariant representation to learn low granularity general features, at this stage training is supervised by pseudo-labels allowing to learn medium granularity features associated with class information. In fact, one model of the ensemble is chosen to be trained on original pseudo-labeled samples. Our criteria for choosing a block from the ensemble is the overall error over initial testing on the pseudo-labeled data. Specifically for each sample $X$ with label $Y$, the label is the same size as the output of the projector, with at least one and at most ${m}$ non-zero elements (hard and soft labels respectively). The chosen model is trained to learn cluster (pseudo-class) assignments guided by the labels. The intriguing point here is that the assigned labels might not necessarily be the hard label corresponding to the  actual class of the sample. In fact, in this fuzzy annotation, assuming an effective first phase of training, it is expected that the pseudo-labeling process should find cross-class features between cross-class samples which would make an even more effective second training phase than the case of only training with hard labels. As a core idea in problem domains such as continual learning \cite{ren2018cross,soutif2021importance,ebrahimi2019uncertainty,fini2022self,de2021continual} or other domains\cite{bart2005cross}, learning invariant features shared across the tasks or classes (in class incremental setting) enhances the generalizability and robustness. The second phase of training is primarily to learn cross-class features along with the class specific features, noting that each class could be an actual class or a cluster.   

As the type of supervisory signal for the second phase training is different, we expect regularization of formerly trained models that were solely based on rigorous invariant representation. Roughly speaking the former phase of training would be categorized as second order statistical constraint on the representation, whereas the latter phase of training would be considered as first order geometrical constraint on the sample representation in the representation space. Former geometrical baselines such as \cite{caron2020unsupervised} proved to be very robust in multiple settings as they naturally tend to learn higher granularity of features due to the fact that their supervisory signal relies on class/cluster level feature learning as opposed to sample level invariant representation.
\vspace{-.31cm}
\section{Experiments and results}
\vspace{-.15cm}
In this section we present our experiments on re-evaluating multiple baselines pre-trained via FUSSL training protocol. These pre-training are performed on ImageNet dataset \cite{deng2009imagenet}, and evaluated in two settings, linear evaluation on ImageNet as well as transfer learning on CIFAR10 and CIFAR100 \cite{krizhevsky2009learning}, and also ablation study on Tiny ImageNet \cite{le2015tiny}. Thanks to newly released Solo-Learn library \cite{da2022solo} we perform all experiments for the first phase of training using Solo-Learn, whereas the second phase is performed also mainly assisted by the their open access code. 

\textbf{Datasets:} CIFAR10 is a dataset composed of 60k images of dimensions $32\times\ 32$  in 10 classes, with 50k images for training and 10k for testing. CIFAR100 is the same size as CIFAR10 and the same sample dimensions, except in 100 classes and also 20 superclasses. Each sample has a fine label (class label) as well as a course label (superclass label). ImageNet is a large scale dataset with multiple versions, where the most common version is composed of over 1.2 million training images and 100k test  images of dimensions $224\times224$ all in 1k classes. Tiny ImageNet is also a smaller version of ImageNet with over 100k samples of dimension $64 \times 64$ in 200 classes.

\textbf{Baselines:} Baselines include one contrastive, two non-contrastive, one geometrical, and two whitening (redundancy reduction) baselines, namely, SimCLR \cite{chen2020simple}, BYOL \cite{grill2020bootstrap}, SimSiam \cite{chen2021exploring}, SwAV \cite{caron2020unsupervised}, Whitening-MSE ($d=4$) \cite{ermolov2021whitening} and Barlow Twins \cite{zbontar2021barlow}.
For SimCLR as a contrastive baseline we followed the original formulation \cite{chen2020simple} with $\tau=0.5$. Following the original implementation of B-Twins, we set $\lambda=5\times 10^{-3}$. SwAV is a clustering based method representing very robust results in a number of settings. Similar to prior  work such as \cite{ermolov2021whitening} we nort-2 normalize the latent space in all baselines. 
\vspace{-0.21cm}
\subsection{Experimental setting}
\vspace{-0.15cm}
\subsubsection{Architecture:} 
Following the details in the above mentioned baselines \cite{chen2020simple,grill2020bootstrap,chen2021exploring,caron2020unsupervised,ermolov2021whitening,zbontar2021barlow}, we use ResNet50 \cite{he2016deep} as the base architecture for the backbone for all experiments performed for any baseline except that last layer is substituted with a projector head with linear layers. The projector is composed of two consecutive layers each followed by batch norm and ReLU, and a third layer as the output, all of size 1000 \cite{zbontar2021barlow}. The size of projector output is identical in all baselines. 
\vspace{-.21cm}
\subsubsection{Augmentation protocol:} As mentioned, for a given sample $x$ from ImageNet and ensemble-${m}$, the augmentation process is tasked to provide $2{m}$ augmented views and allocate a distinct pair to each block of ensemble. That said, the augmentation over phase one pre-training is executed similar to all former baselines under a random distribution $\tau$ which enables randomness in a set of augmentation techniques including random crop, random color jittering, mirroring, and random aspect ratio specification with exact settings suggested by \cite{chen2020simple}. ImageNet images with size $224 \times 224$ undergo a random crop of size between $0.2$ and whole image size, an aspect ratio adjustment anywhere randomly chosen between $3/4$ and $4/3$, and random horizontal mirroring distributed with mean $1/2$  and finally color jittering anywhere randomly chosen between the spectra  $(0.4, 0.4, 0.4, 0.1)$ and graysaling with probability ratio of 8 to 1. However for second phase of training and evaluation we performed no augmentation. Note that the main scenario of training sample allocation is that each block only sees its pair of augmented views for a given sample, aiming for uncertainty injection to the training process.
\vspace{-.15cm}
\subsubsection{Implementation details:} 
\label{sec413}
For all experiments including ablation study the optimisation for both pre-training on ImageNet or Tiny ImageNet and training session of the evaluation on ImageNet, Tiny ImageNet and CIFAR10/100 is performed using the Adam optimizer \cite{kingma2014adam}. We followed the settings presented in \cite{chen2020simple} for transfer learning using pre-trained ResNet50 on CIFAR10/100 performed for all six baselines.\\
\textbf{Phase 1 training:} First phase of training is performed on an ensemble-${m}$ of a given baseline with size ${m}=3$, which involves 800 epochs of training with batch size of 1024, which initiates with a learning rate of $0.2$ for some 10 epochs and drops to $0.001$ for the remaining epochs. The weight decay is $10^{-6}$ for all experiments including corresponding experiments presented in ablation study. We examine the results for other size of ${m}$ later in section 5.\\
\textbf{Fuzzy pseudo-labeling:} Upon the completion of phase 1 training, we freeze the weight of both backbone and projector architectures of all blocks of ensemble, and examine the normalized output for each original sample. Hence for each sample, there would be ${m}$ outputs each with size 1000 as the pseudo-classes (clusters). Considering the largest number in the output vector as the pseudo-class assigned by a block, the most frequent pseudo-class is set as the hard label of the sample, otherwise in case of ${m}$ different pseudo-classes, a soft label would be assigned to the sample. Soft labels are to target cross-class features while hard labels are to target class specific features for the next phase of training.\\ 
\textbf{Phase 2 training:} Second phase of training is guided by another type of supervisory signal, i.e., pseudo-labels, aimed at learning medium to high granularity features. The chosen model from one of the blocks is trained on pseudo-labeled samples for 400 epochs with a learning rate of $0.001$, where first 100 epochs back-propagate the error on both backbone and projector whereas for remaining 300 epochs, only last stage of ResNet50 (last 9 layers) along with the projector head are trained, i.e., we fix first 41 layers. Therefore, the training during the remaining 300 epochs is conditioned on weights of those fixed layers. Another scenario with no fixed weights is considered in ablation study.
\vspace{-0.3cm}
\subsection{Evaluation}
\vspace{-0.2cm}
In this section,
we investigate the effectiveness of the FUSSL training protocol, we present the the evaluation of the baselines, with and without FUSSL protocol via the standard practice \cite{goyal2019scaling} for classification tasks. This is technically to evaluate the baselines right after the first phase of training as well as after second phase of training.  
The most common standard procedure for evaluating the SSL pre-training techniques is to train and test a classifier on top of the backbone architecture with fixed weights; even though less common approach is using K-nearest neighbourhood classifier without further training. 
In this work, after first phase and also second phase of pre-training using FUSSL, all six SSL baselines are to be evaluated under the standard protocol for training a linear classifier (e.g., a fully connected layer and a softmax) on the downstream task labeled data. We follow the details of the most recent baseline \cite{ermolov2021whitening}, in both phases of evaluation we train the classifier for some 500 epochs and then test it.
Evaluation is performed in two different settings, linear evaluation as well as transfer learning. Linear evaluation only involves ImageNet dataset, whereas transfer learning is performed on CIFAR10 and CIFAR100. In both cases the pre-training phases is performed on ImageNet samples. Similar to standard linear evaluation, in case of transfer learning, the fixed per-trained backbone followed by a linear classifier (fully connected followed by softmax) will be trained for some 500 epochs on the dataset, e.g., CIFAR10 or CIFAR100, and tested.
The number of both phases of pre-training epochs as well as evaluation epochs are carefully selected, however further experiments are presented in the ablation study.
\vspace{-0.3cm}
\subsection{Results}
\vspace{-0.2cm}
\textbf{Linear evaluation:} Table 1 
presents the results for linear evaluation, top-1 classification accuracy on ImageNet, pre-trained on the same dataset (with no labels). As it is shown, except for the SwAV as a clustering approach, FUSSL noticeably improves the accuracy of other five baselines ranging from $0.8\%$ (BYOL) to $1.4\%$ (B-Twins) improvement. We suspect that the reason that FUSSL improves SwAV by only $0.3\%$ is that SwAV as a geometrical approach already enjoys the medium to high granularity feature learning via clustering techniques. Hence, the double supervision signal used in FUSSL does not provide much advantage for SwAV.
\begin{table}
  \label{table1}
  \centering
  \scriptsize
  \begin{tabular}{p{1.4cm} p{1.5cm}|p{1.5cm}| p{2.5cm}}
    \toprule
    Framework   & \multicolumn{3}{c}{ImageNet}   \\
    \cmidrule(r){2-4}
        & Base   & 1200 ep& FUSSL \\
    \cmidrule(r){1-4}
     SimCLR  & 69.3 &  69.4 &   70.4 \: ($1.1\% \uparrow$ ) \\
      BYOL  &  73.9  & 74.1&    74.7 \: ($0.8\% \uparrow $ ) \\
       SwAV  &  75.1  & 75.2 & 75.4  \: ($0.3\% \uparrow $ )\\
       SimSiam  &  70.9 & 70.8 &  72.1  \: ($1.2\% \uparrow $ )  \\
    W-MSE4  &  73.1 & 73.4&  74.4 \: ($1.3\% \uparrow $ )\\
   \textbf{B-Twins}  &   \textbf{73.3} &{73.4}  & \textbf{74.7} \: (\textbf{1.4}$\% \uparrow $ ) \\
    \bottomrule
  \end{tabular}
  \caption{
  \footnotesize
  Top-1 linear classification accuracy for ImageNet using ResNet50 pre-trained on ImageNet under three settings. "Base" and "1200 ep" represent the results of evaluation of each baseline under 800 and 1200 epochs of pre-training respectively; whereas "FUSSL" shows the results for the same baselines under two phases of pre-training, i.e., 800 epochs for phase 1 of pre-training and 400 epochs for phase 2 (total of 1200 epochs).}
  \vspace{-0.3cm}
\end{table}

\begin{table}
  \label{table2}
  \centering
  \scriptsize
  \begin{tabular}{p{1.cm} p{0.8cm} p{0.8cm} p{0.8cm}|p{0.8cm} p{0.8cm} p{0.8cm}}
    \toprule
    Framework   & \multicolumn{3}{c}{CIFAR10}     & \multicolumn{3}{c}{CIFAR100} \\
    \cmidrule(r){2-7}
        & Base &1200 ep  & FUSSL &  Base & 1200 ep & FUSSL \\
    \cmidrule(r){2-7}
     SimCLR  & 89.97 &90.11 & 90.83  & 75.91   & 75.99 &    76.59\\
      BYOL  &   91.28  & 91.42& 92.11  &  78.49 & 78.75& 79.53   \\
       SwAV  &   94.33 & 94.41& 94.78 &  81.01 & 81.11& 81.33 \\
       SimSiam  &    92.75 & 92.97& 94.02  & 78.36  & 78.48 &  79.21 \\
    \textbf{W-MSE4}  & \textbf{94.88} &  {95.12}  & \textbf{96.15}    & 79.01     & 79.23&  80.15 \\
   \textbf{B-Twins}  &    95.12 & 95.40& 96.33 &  \textbf{80.18}    &  80.35  &     \textbf{81.40}    \\
    \bottomrule
  \end{tabular}
  \caption{
  \footnotesize
  Top-1 transfer learning classification accuracy for CIFAR10/100 using ResNet50 pre-trained on ImageNet under three settings.
}
\vspace{-0.3cm}
\end{table}
To emphasize  the significance of the results, it is important to mention that except for BYOL as a breakthrough baseline and to a lesser degree SwAV, most of the recent SSL baselines either do not offer accuracy improvement or offer less than $1\%$ improvement over former baselines under a full pre-training of 400 epochs or more. For instance, SimSiam is a baseline introduced after BYOL, with the advantage of faster learning convergence (highest accuracy in 100 epochs compared to former baselines) and with no improvement over former baselines under long pre-training (400 epochs or more) on ImageNet. The same behavior is noticeable for W-MSE or B-Twins, both of which under-performed  their predecessor baseline, BYOL for a full pre-training of 400 epochs or more. However, FUSSL consistently improves the performance of each baseline, thanks to its innovation in using double supervisory signals.

\textbf{Transfer learning:} The results of transfer learning are shown in Table 2,
for two datasets, CIFAR10/100. The most  and least improvements were consistently observed in 
B-Twins and SwAV, respectively. In this setting,  FUSSL also improves the performance of each baseline.  
\vspace{-0.21cm}
\section{Ablation study}
\begin{figure}
    \centering
   \includegraphics[scale=0.091]{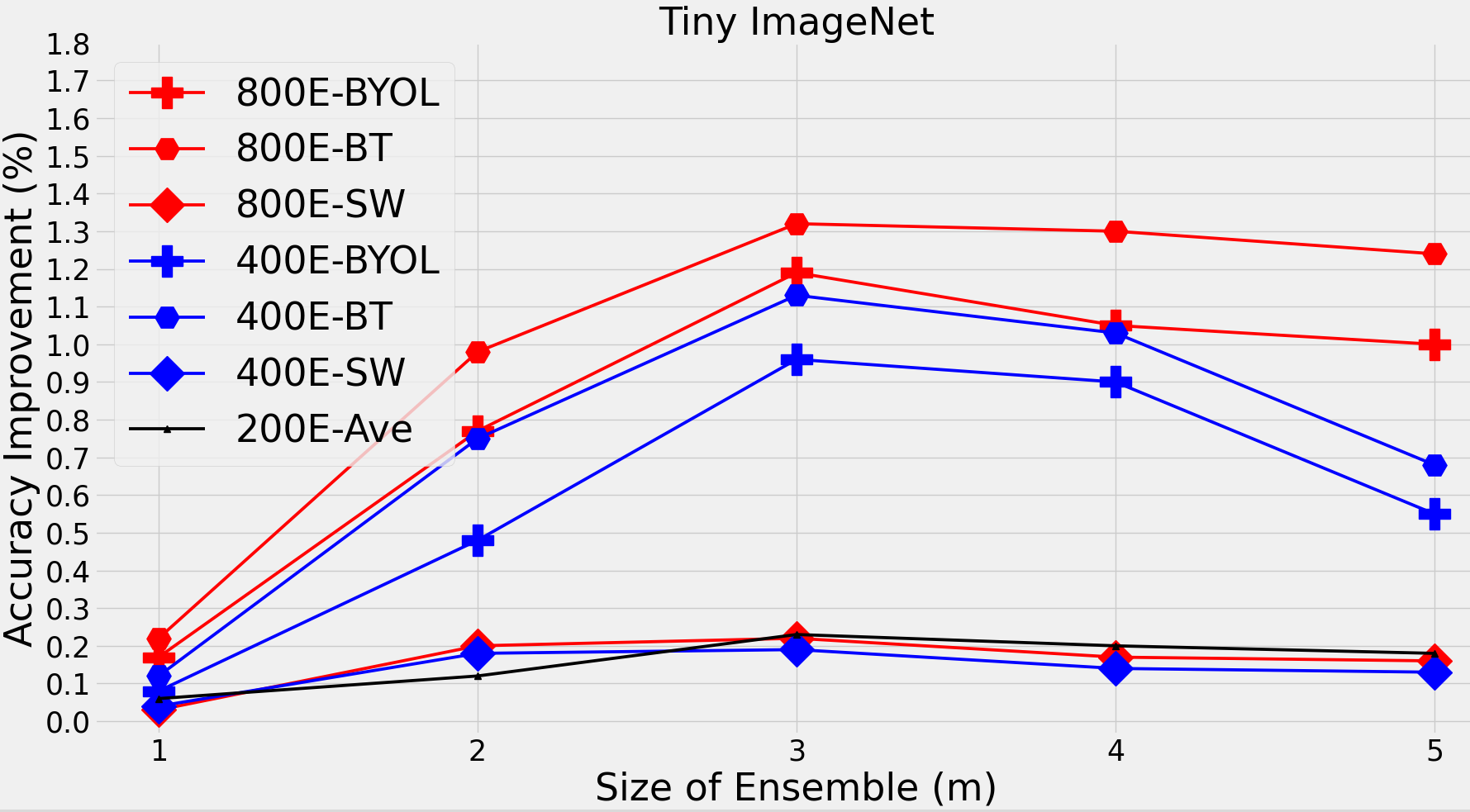}
    \caption{\scriptsize
    Individual and average accuracy  improvement for SwAV, BOL and B-Twins, under different ensemble sizes (computational cost), and using 800 and 400, as well as 200 epochs of first phase pre-training. Best result with $m=3$.}
    \label{fig.1}
    \vspace{-0.3cm}
\end{figure}

\vspace{-0.15cm}
In this section we analyze the components of the proposed training protocol under multiple scenarios. These scenarios include hard vs soft labeling, size of ensemble, progressive pseudo-labeling, and cross-class/cluster feature learning. We also briefly consider the resilience of the model to backdoor attack on SSL as characterized in \cite{saha2022backdoor}. Except for cross-class/cluster feature learning which is performed on CIFAR100, all other scenarios are assessed on Tiny ImageNet \cite{le2015tiny} for examples of three categories of baselines: clustering-based (SwAV), non-contrastive (BYOL), and  whitening (B-Twins) baselines. 
Size of ensemble is 3 unless otherwise stated.\\
\indent \textbf{Hard vs soft:} Case 1 and Case 2 in Table 3 (Sec. \ref{table3}) present the results of FUSSL using barely hard or soft labels, respectively. It seems that FUSSL results are better owing to the soft labeling, as soft labeling provides improved results when compared to  hard labeling.\\
\indent \textbf{Size of ensemble:} We assess the results for ensemble-${m}$ with five different $m$ values. Fig. \ref{fig.1} shows the results of individual and average improvement over three baselines under different number of epochs for first phase training, as $m$ directly involves the first phase. Note that in the case of ensemble with size $m=1$, the ensemble is essentially one block, and the pseudo-labeling reduces to hard labeling as we only access the  output of one block. As shown in Fig. \ref{fig.1}, $m=3$ delivers the highest average improvement.\\
\indent \textbf{Progressive pseudo-labeling:} A very interesting scenario, is to evaluate the case in which a series of first and second phase training is performed. More specifically, after each 200 epochs of phase 1, pseudo-labeling is performed and applied to  100 epochs of phase 2, and so on. In each iteration of (phase 1) to (phase 2), there is  no fixed layer and pseudo-labeling is performed from scratch. As shown in Table 3 
Case 3, the average improvement over three baselines drops to $0.37\%$, which 
underscores the effectiveness of our 
strategy (Sec \ref{sec3intro}) to minimize the mutual information between learned features from the two learning phases.\\
\indent \textbf{Cross-class/cluster learning:} Inspired by \cite{lienen2021credal,fini2022self}, in this scenario we assess the effectiveness of hard and soft pseudo-labeling for cross-class feature learning. We explore the idea that assigning more than one label (here $m$ labels) to a sample, essentially enables learning the features that are shared between those $m$ classes. To this end, considering that CIFAR100 comes in 100 classes and 20 superclasses, we perform the pre-training on only half of the samples across all superclasses and only from 50 classes, and then we evaluate the performance over the other 50 classes (still in all 20 superclasses) in two settings. In other words, pre-training data and evaluation data are from different classes but the same superclassess. We suspect 
that soft labeling would help the model to learn the general features of a given superclass via pre-training on half of its classes, so that evaluation on other half of the classes benefits from the cross-class learned features. As shown in Table 4, 
Case 4 represents only using the hard labels whereas Case 5 represents only assigning soft labels. 
Table 4 show that using soft labeling significantly improves the performance. (See supplementary material for more details).\\
\indent \textbf{No fixed weights:} As mentioned in Sec. \ref{sec413}, during the second phase of training, immediately after first 100 epochs, first 41 layers of backbone ResNet50 are fixed. However, we also examined the case in which all 50 layers are being trained (no fixed layer). In case of retraining all layers of ResNet50, the average improvements over three baselines drops from $0.92\%$ to $0.28\%$ which clearly indicates the effectiveness of our practical strategy of conditional learning mentioned in introduction of Sec. \ref{sec3intro}.\\
\indent \textbf{Backdoor attack:} We also assessed the robustness of the protocol in the context of backdoor attack to SSL as recently introduced in \cite{saha2022backdoor}, where adding a small amount of unhealthy data during the pre-training, fools the model
at the downstream tasks. 
Our results (not shown) indicate how FUSSL with $m>1$ under careful settings improves the robustness of a given baseline against backdoor attack.\\
\indent \textbf{Limitation:} One  limitation of this work is the relatively low improvement for cluster-based approaches, e.g., SwAV,
possibly due to the type of supervision which enables them to learn medium to high granularity  features which generally benefits the downstream task. In fact, compared to sample-level, cluster-level learning tends to offer features often associated 
with higher level semantic meanings.  
\begin{table}
  \label{table3}
  \centering
  \scriptsize
  \begin{tabular}{p{01.3cm} p{0.9cm}| p{0.9cm}| p{01cm}| p{1cm}| p{1cm}}
    \toprule
    Framework   & \multicolumn{4}{c}{Tiny ImageNet}   \\
    \cmidrule(r){2-6}
        & Base   & FUSSL & Case 1 & Case 2 & Case 3\\
    \cmidrule(r){1-6}
      BYOL  &   51.45 &   52.54   &  51.86   &  52.16   & 51.84\\
       SwAV  & 51.60 &   51.93   &  51.66    &  51.80     & 51.81\\
   B-Twins  &   50.89  &   52.21   &    51.61  &    51.91  &  51.39\\
    \bottomrule
  \end{tabular}
  \caption{\footnotesize
  Cases 1 and 2 represent the results for hard labeling and soft labeling respectively, Case 3 represents the results for progressive pseudo-labeling (Base column: results of the baselines without FUSSL protocol
  ). }
\end{table}

  \begin{table}
  \label{table4}
  \centering
  \scriptsize
  \begin{tabular}{p{01.3cm} p{0.9cm}|p{0.9cm}| p{0.9cm}| p{01cm}}
    \toprule
    Framework   & \multicolumn{4}{c}{CIFAR100}   \\
    \cmidrule(r){2-5}
        & Base   & FUSSL    &   Case 4 & Case 5 \\
    \cmidrule(r){1-5}
      BYOL  &   62.93  & 63.97 & 63.20  & 63.81  \\
       SwAV  &  63.15 & 63.47 & 63.27     & 63.32 \\
   B-Twins  &   62.89 &  64.19 &  63.11  & 64.02\\
    \bottomrule
  \end{tabular}
  \caption{\footnotesize
  Cross-class/cluster feature learning, Case 4 represents the results for hard labeling, whereas Case 5 represents the results for soft labeling. Note that Base, FUSSL, Cases 4 and 5 are all pre-trained on 50 classes and evaluated on remaining 50 classes of CIFAR100. The results are in line 
  with our expectation that, assigning soft labels (here $m$ labels, $m=$size of ensemble) to the samples allows learning features of the superclasses, which is shared across the assigned pseudo-classes. In essence, this is sort of transfer learning between classes belonging to the same superclass. 
  }
\vspace{-0.3cm}
\end{table}
\vspace{-0.3cm}
\section{Conclusion}
\vspace{-0.24cm}
In this paper, we investigate recent SSL baselines from the perspective of self-supervisory signals, and identify the pros and cons of 
each type of supervision. Accordingly, building on the information bottleneck principle, we propose FUSSL as a general training protocol for SSL frameworks based on sample-level as well as cluster-level feature learning, enabling the learning of features via two self-supervisory signals. First supervisory signal guides the first phase, followed by fuzzy pseudo-labeling and then second phase of learning. We provided a foundation and its translation into experimental settings to ensure consistent improvement over all former baselines. Extensive experiments and detailed ablation studies 
suggest the effectiveness of the proposed protocol for SSL frameworks, regardless of the specific approach.

\textbf{Acknowledgement}\\
This work is supported in part by grants from the US National Science
Foundation (Award $\#1920920$, $\#2125872$).

{\small
\bibliographystyle{ieee_fullname}
\bibliography{egbib}

\begin{thebibliography}{10}\itemsep=-1pt

\bibitem{Lexpodcast}
Lex fridman podcast, ({MIT AI} podcast), episode \# 258: Dark matter of
  intelligence and self-supervised learning , 2022.

\bibitem{bachman2019learning}
Philip Bachman, R~Devon Hjelm, and William Buchwalter.
\newblock Learning representations by maximizing mutual information across
  views.
\newblock {\em Advances in Neural Information Processing Systems}, 32, 2019.

\bibitem{balestriero2022contrastive}
Randall Balestriero and Yann LeCun.
\newblock Contrastive and non-contrastive self-supervised learning recover
  global and local spectral embedding methods.
\newblock {\em arXiv preprint arXiv:2205.11508}, 2022.

\bibitem{bart2005cross}
Evgeniy Bart and Shimon Ullman.
\newblock Cross-generalization: Learning novel classes from a single example by
  feature replacement.
\newblock In {\em 2005 IEEE Computer Society Conference on Computer Vision and
  Pattern Recognition (CVPR'05)}, volume~1, pages 672--679. IEEE, 2005.

\bibitem{bautista2016cliquecnn}
Miguel~A Bautista, Artsiom Sanakoyeu, Ekaterina Tikhoncheva, and Bjorn Ommer.
\newblock Clique{CNN}: Deep unsupervised exemplar learning.
\newblock {\em Advances in Neural Information Processing Systems}, 29, 2016.

\bibitem{becker1992self}
Suzanna Becker and Geoffrey~E Hinton.
\newblock Self-organizing neural network that discovers surfaces in random-dot
  stereograms.
\newblock {\em Nature}, 355(6356):161--163, 1992.

\bibitem{caron2018deep}
Mathilde Caron, Piotr Bojanowski, Armand Joulin, and Matthijs Douze.
\newblock Deep clustering for unsupervised learning of visual features.
\newblock In {\em Proceedings of the European Conference on Computer Vision
  (ECCV)}, pages 132--149, 2018.

\bibitem{caron2020unsupervised}
Mathilde Caron, Ishan Misra, Julien Mairal, Priya Goyal, Piotr Bojanowski, and
  Armand Joulin.
\newblock Unsupervised learning of visual features by contrasting cluster
  assignments.
\newblock {\em Advances in Neural Information Processing Systems},
  33:9912--9924, 2020.

\bibitem{caron2021emerging}
Mathilde Caron, Hugo Touvron, Ishan Misra, Herv{\'e} J{\'e}gou, Julien Mairal,
  Piotr Bojanowski, and Armand Joulin.
\newblock Emerging properties in self-supervised vision transformers.
\newblock In {\em Proceedings of the IEEE/CVF International Conference on
  Computer Vision}, pages 9650--9660, 2021.

\bibitem{chen2020simple}
Ting Chen, Simon Kornblith, Mohammad Norouzi, and Geoffrey Hinton.
\newblock A simple framework for contrastive learning of visual
  representations.
\newblock In {\em International Conference on Machine Learning}, pages
  1597--1607. PMLR, 2020.

\bibitem{chen2021exploring}
Xinlei Chen and Kaiming He.
\newblock Exploring simple siamese representation learning.
\newblock In {\em Proceedings of the IEEE/CVF Conference on Computer Vision and
  Pattern Recognition}, pages 15750--15758, 2021.

\bibitem{da2022solo}
Victor Guilherme~Turrisi da Costa, Enrico Fini, Moin Nabi, Nicu Sebe, and Elisa
  Ricci.
\newblock solo-learn: A library of self-supervised methods for visual
  representation learning.
\newblock {\em J. Mach. Learn. Res.}, 23:56--1, 2022.

\bibitem{de2021continual}
Matthias De~Lange, Rahaf Aljundi, Marc Masana, Sarah Parisot, Xu Jia,
  Ale{\v{s}} Leonardis, Gregory Slabaugh, and Tinne Tuytelaars.
\newblock A continual learning survey: Defying forgetting in classification
  tasks.
\newblock {\em IEEE Transactions on Pattern Analysis and Machine Intelligence},
  44(7):3366--3385, 2021.

\bibitem{deng2009imagenet}
Jia Deng, Wei Dong, Richard Socher, Li-Jia Li, Kai Li, and Li Fei-Fei.
\newblock Imagenet: A large-scale hierarchical image database.
\newblock In {\em 2009 IEEE Conference on Computer Vision and Pattern
  Recognition}, pages 248--255. Ieee, 2009.

\bibitem{ebrahimi2019uncertainty}
Sayna Ebrahimi, Mohamed Elhoseiny, Trevor Darrell, and Marcus Rohrbach.
\newblock Uncertainty-guided continual learning with bayesian neural networks.
\newblock {\em arXiv preprint arXiv:1906.02425}, 2019.

\bibitem{ermolov2021whitening}
Aleksandr Ermolov, Aliaksandr Siarohin, Enver Sangineto, and Nicu Sebe.
\newblock Whitening for self-supervised representation learning.
\newblock In {\em International Conference on Machine Learning}, pages
  3015--3024. PMLR, 2021.

\bibitem{fini2022self}
Enrico Fini, Victor G~Turrisi da Costa, Xavier Alameda-Pineda, Elisa Ricci,
  Karteek Alahari, and Julien Mairal.
\newblock Self-supervised models are continual learners.
\newblock In {\em Proceedings of the IEEE/CVF Conference on Computer Vision and
  Pattern Recognition}, pages 9621--9630, 2022.

\bibitem{goyal2019scaling}
Priya Goyal, Dhruv Mahajan, Abhinav Gupta, and Ishan Misra.
\newblock Scaling and benchmarking self-supervised visual representation
  learning.
\newblock In {\em Proceedings of the ieee/cvf International Conference on
  Computer Vision}, pages 6391--6400, 2019.

\bibitem{grill2020bootstrap}
Jean-Bastien Grill, Florian Strub, Florent Altch{\'e}, Corentin Tallec, Pierre
  Richemond, Elena Buchatskaya, Carl Doersch, Bernardo Avila~Pires, Zhaohan
  Guo, Mohammad Gheshlaghi~Azar, et~al.
\newblock Bootstrap your own latent-a new approach to self-supervised learning.
\newblock {\em Advances in Neural Information Processing Systems},
  33:21271--21284, 2020.

\bibitem{he2020momentum}
Kaiming He, Haoqi Fan, Yuxin Wu, Saining Xie, and Ross Girshick.
\newblock Momentum contrast for unsupervised visual representation learning.
\newblock In {\em Proceedings of the IEEE/CVF Conference on Computer Vision and
  Pattern Recognition}, pages 9729--9738, 2020.

\bibitem{he2016deep}
Kaiming He, Xiangyu Zhang, Shaoqing Ren, and Jian Sun.
\newblock Deep residual learning for image recognition.
\newblock In {\em Proceedings of the IEEE Conference on Computer Vision and
  Pattern Recognition}, pages 770--778, 2016.

\bibitem{hendrycks2019using}
Dan Hendrycks, Mantas Mazeika, Saurav Kadavath, and Dawn Song.
\newblock Using self-supervised learning can improve model robustness and
  uncertainty.
\newblock {\em Advances in Neural Information Processing Systems}, 32, 2019.

\bibitem{huang2022learning}
Lang Huang, Shan You, Mingkai Zheng, Fei Wang, Chen Qian, and Toshihiko
  Yamasaki.
\newblock Learning where to learn in cross-view self-supervised learning.
\newblock In {\em Proceedings of the IEEE/CVF Conference on Computer Vision and
  Pattern Recognition}, pages 14451--14460, 2022.

\bibitem{ji2019invariant}
Xu Ji, Joao~F Henriques, and Andrea Vedaldi.
\newblock Invariant information clustering for unsupervised image
  classification and segmentation.
\newblock In {\em Proceedings of the IEEE/CVF International Conference on
  Computer Vision}, pages 9865--9874, 2019.

\bibitem{jing2020self}
Longlong Jing and Yingli Tian.
\newblock Self-supervised visual feature learning with deep neural networks: A
  survey.
\newblock {\em IEEE Transactions on Pattern Analysis and Machine Intelligence},
  43(11):4037--4058, 2020.

\bibitem{kingma2014adam}
Diederik~P Kingma and Jimmy Ba.
\newblock Adam: A method for stochastic optimization.
\newblock {\em arXiv preprint arXiv:1412.6980}, 2014.

\bibitem{krizhevsky2009learning}
Alex Krizhevsky, Geoffrey Hinton, et~al.
\newblock Learning multiple layers of features from tiny images.
\newblock 2009.

\bibitem{le2015tiny}
Ya Le and Xuan Yang.
\newblock Tiny imagenet visual recognition challenge.
\newblock {\em CS 231N}, 7(7):3, 2015.

\bibitem{lienen2021credal}
Julian Lienen and Eyke H{\"u}llermeier.
\newblock Credal self-supervised learning.
\newblock {\em Advances in Neural Information Processing Systems},
  34:14370--14382, 2021.

\bibitem{liu2019exploiting}
Xialei Liu, Joost Van De~Weijer, and Andrew~D Bagdanov.
\newblock Exploiting unlabeled data in cnns by self-supervised learning to
  rank.
\newblock {\em IEEE Transactions on Pattern Analysis and Machine Intelligence},
  41(8):1862--1878, 2019.

\bibitem{misra2016shuffle}
Ishan Misra, C~Lawrence Zitnick, and Martial Hebert.
\newblock Shuffle and learn: unsupervised learning using temporal order
  verification.
\newblock In {\em European Conference on Computer Vision}, pages 527--544.
  Springer, 2016.

\bibitem{mohamadi2020deep}
Salman Mohamadi and Hamidreza Amindavar.
\newblock Deep bayesian active learning, a brief survey on recent advances.
\newblock {\em arXiv preprint arXiv:2012.08044}, 2020.

\bibitem{mohamadi2022deep}
Salman Mohamadi, Gianfranco Doretto, and Donald~A Adjeroh.
\newblock Deep active ensemble sampling for image classification.
\newblock {\em arXiv preprint arXiv:2210.05770}, 2022.

\bibitem{motiian2016information}
Saeid Motiian, Marco Piccirilli, Donald~A Adjeroh, and Gianfranco Doretto.
\newblock Information bottleneck learning using privileged information for
  visual recognition.
\newblock In {\em Proceedings of the IEEE Conference on Computer Vision and
  Pattern Recognition}, pages 1496--1505, 2016.

\bibitem{nava2021uncertainty}
Mirko Nava, Antonio Paolillo, J{\'e}r{\^o}me Guzzi, Luca~Maria Gambardella, and
  Alessandro Giusti.
\newblock Uncertainty-aware self-supervised learning of spatial perception
  tasks.
\newblock {\em IEEE Robotics and Automation Letters}, 6(4):6693--6700, 2021.

\bibitem{oord2018representation}
Aaron van~den Oord, Yazhe Li, and Oriol Vinyals.
\newblock Representation learning with contrastive predictive coding.
\newblock {\em arXiv preprint arXiv:1807.03748}, 2018.

\bibitem{poggi2020uncertainty}
Matteo Poggi, Filippo Aleotti, Fabio Tosi, and Stefano Mattoccia.
\newblock On the uncertainty of self-supervised monocular depth estimation.
\newblock In {\em Proceedings of the IEEE/CVF Conference on Computer Vision and
  Pattern Recognition}, pages 3227--3237, 2020.

\bibitem{ren2018cross}
Zhongzheng Ren and Yong~Jae Lee.
\newblock Cross-domain self-supervised multi-task feature learning using
  synthetic imagery.
\newblock In {\em Proceedings of the IEEE Conference on Computer Vision and
  Pattern Recognition}, pages 762--771, 2018.

\bibitem{roh2021spatially}
Byungseok Roh, Wuhyun Shin, Ildoo Kim, and Sungwoong Kim.
\newblock Spatially consistent representation learning.
\newblock In {\em Proceedings of the IEEE/CVF Conference on Computer Vision and
  Pattern Recognition}, pages 1144--1153, 2021.

\bibitem{saha2022backdoor}
Aniruddha Saha, Ajinkya Tejankar, Soroush~Abbasi Koohpayegani, and Hamed
  Pirsiavash.
\newblock Backdoor attacks on self-supervised learning.
\newblock In {\em Proceedings of the IEEE/CVF Conference on Computer Vision and
  Pattern Recognition}, pages 13337--13346, 2022.

\bibitem{sohn2016improved}
Kihyuk Sohn.
\newblock Improved deep metric learning with multi-class n-pair loss objective.
\newblock {\em Advances in Neural Information Processing Systems}, 29, 2016.

\bibitem{soutif2021importance}
Albin Soutif-Cormerais, Marc Masana, Joost Van~de Weijer, and Bartl{\o}miej
  Twardowski.
\newblock On the importance of cross-task features for class-incremental
  learning.
\newblock {\em CoRR}, 2021.

\bibitem{tian2021understanding}
Yuandong Tian, Xinlei Chen, and Surya Ganguli.
\newblock Understanding self-supervised learning dynamics without contrastive
  pairs.
\newblock In {\em International Conference on Machine Learning}, pages
  10268--10278. PMLR, 2021.

\bibitem{tian2020contrastive}
Yonglong Tian, Dilip Krishnan, and Phillip Isola.
\newblock Contrastive multiview coding.
\newblock In {\em European Conference on Computer Vision}, pages 776--794.
  Springer, 2020.

\bibitem{tishby2000information}
Naftali Tishby, Fernando~C Pereira, and William Bialek.
\newblock The information bottleneck method.
\newblock {\em arXiv preprint physics/0004057}, 2000.

\bibitem{wang2015unsupervised}
Xiaolong Wang and Abhinav Gupta.
\newblock Unsupervised learning of visual representations using videos.
\newblock In {\em Proceedings of the IEEE International Conference on Computer
  Vision}, pages 2794--2802, 2015.

\bibitem{wang2021dense}
Xinlong Wang, Rufeng Zhang, Chunhua Shen, Tao Kong, and Lei Li.
\newblock Dense contrastive learning for self-supervised visual pre-training.
\newblock In {\em Proceedings of the IEEE/CVF Conference on Computer Vision and
  Pattern Recognition}, pages 3024--3033, 2021.

\bibitem{xiao2021region}
Tete Xiao, Colorado~J Reed, Xiaolong Wang, Kurt Keutzer, and Trevor Darrell.
\newblock Region similarity representation learning.
\newblock In {\em Proceedings of the IEEE/CVF International Conference on
  Computer Vision}, pages 10539--10548, 2021.

\bibitem{xie2021propagate}
Zhenda Xie, Yutong Lin, Zheng Zhang, Yue Cao, Stephen Lin, and Han Hu.
\newblock Propagate yourself: Exploring pixel-level consistency for
  unsupervised visual representation learning.
\newblock In {\em Proceedings of the IEEE/CVF Conference on Computer Vision and
  Pattern Recognition}, pages 16684--16693, 2021.

\bibitem{zbontar2021barlow}
Jure Zbontar, Li Jing, Ishan Misra, Yann LeCun, and St{\'e}phane Deny.
\newblock Barlow twins: Self-supervised learning via redundancy reduction.
\newblock In {\em International Conference on Machine Learning}, pages
  12310--12320. PMLR, 2021.

\end{thebibliography}
}

\end{document}